\documentclass[preprint,12pt]{elsarticle}
\usepackage[T1]{fontenc}
\usepackage[utf8]{inputenc}
\usepackage{amsmath,amssymb,booktabs,tabularx}
\usepackage{microtype,placeins,xurl,etoolbox,needspace}
\usepackage[hidelinks]{hyperref}
\biboptions{numbers,sort&compress}
\newcommand{\clip}{\operatorname{clip}}
\newcommand{\med}{\operatorname{median}}
\journal{Energy}
\makeatletter
\patchcmd{\ps@pprintTitle}{Preprint submitted to}{Manuscript draft for}{}{}
\makeatother
\begin{document}
\begin{frontmatter}
\title{Horizon-specific expert fusion for photovoltaic power forecasting}
\author[1]{Xu Yuqing}
\author[1]{Zhou Liguo}
\author[1]{Sun Ze}
\author[1]{Yu Lei}
\author[1]{Jiang Mingming}
\affiliation[a]{organization={School of Computer Science and Technology, Huaibei Normal University},country={China}}
\begin{abstract}
Short-term photovoltaic power forecasting requires models to represent regular solar cycles and weather-driven fluctuations whose importance changes with the forecast horizon. This study develops a hierarchical ensemble that combines temporal neural models, historical analogs, state climatology, and gradient-boosted trees. Solar geometry and numerical weather forecasts describe the expected generation conditions, while horizon-specific convex weights combine complementary predictions. A separate calibration step uses available historical forecast errors to account for recent bias. The framework is evaluated on public PVDAQ data at 15--240-minute horizons and on three GEFCom2014 solar zones at hourly horizons up to four hours. On PVDAQ, the ensemble achieves a daylight capacity-normalized mean absolute error of 4.315\%, reducing error by 4.11\% relative to full-feature LightGBM and by 6.03\% relative to fine-tuned Chronos-2 under identical calibration. Expert-removal experiments identify redundancy within the ensemble. Across three training seeds on GEFCom2014, learned fusion improves upon equal weighting but performs comparably to LightGBM. The results support horizon-specific combination as a useful forecasting strategy while showing that its advantage over strong individual models depends on the dataset and evaluation period.
\end{abstract}
\begin{keyword}
Photovoltaic forecasting \sep Forecast combination \sep Solar geometry \sep Numerical weather prediction \sep Historical analogs \sep Gradient boosting
\end{keyword}
\end{frontmatter}

\section{Introduction}
Accurate short-term photovoltaic (PV) power forecasts support the scheduling of storage, reserve capacity, and flexible demand. The daily generation envelope follows solar motion, whereas clouds and operating conditions introduce departures from that envelope. These processes act on different time scales. Recent power can be informative over the next few minutes, but its value may decline as the horizon extends and atmospheric conditions change. Reviews of solar and PV forecasting accordingly emphasize the relationship between forecast horizon, available information, and model choice \cite{antonanzas,inman}.

This dependence creates a practical challenge for forecast combination. A model that captures the average daily curve may miss rapid changes, while a model that closely follows recent observations may extend a transient disturbance too far into the future. Numerical weather prediction (NWP) adds information about future atmospheric conditions, but its grid resolution and averaging periods limit its ability to resolve local fluctuations. A useful forecasting system must therefore combine recent measurements and forecast weather without treating either source as uniformly reliable.

Existing methods address different parts of this problem. Multiscale neural models learn temporal structure, gradient-boosted trees capture nonlinear interactions in engineered features, and historical analogs retrieve previously observed generation patterns. Their errors may be complementary, but complementarity cannot be inferred from architecture names alone. Adding weak or redundant predictors can increase both error and computational cost. Stacking has already been applied to distributed PV forecasting, including the framework of Cao et al.\ in \emph{Energy} \cite{cao}. The relevant question is consequently how much a specified fusion strategy improves upon strong individual predictors under a common evaluation protocol.

This study investigates a hierarchical ensemble of eight predictors for forecasts up to four hours ahead. Solar geometry describes diurnal and seasonal position, while archived weather forecasts describe the atmospheric conditions expected at the prediction origin. Three neural predictors, a state-scaled climatology, three rolling analog predictors, and a full-feature LightGBM predictor provide candidate trajectories. Two successive convex combinations assign weights to these trajectories separately for each horizon. A distinct calibration step uses only historical forecast errors whose observations have become available. The fusion weights remain fixed across prediction origins; the system is not presented as a weather-dependent mixture-of-experts gate.

The empirical analysis addresses three questions. Does learned fusion improve upon a full-feature tree model and a forecasting foundation model? Which experts contribute after the remaining weights are refitted? Does the benefit persist in a second public dataset with a different sampling interval and weather interface? We answer these questions using PVDAQ station 2107 and the three solar zones of GEFCom2014. The two experiments retain their native temporal resolutions and are interpreted separately.

The contribution is a specified integration of established forecasting components and a controlled assessment of its conditional value. The study combines complete horizon reporting, common residual calibration, expert removal, repeated training in the hourly adaptation, and component-level timing. This design distinguishes the benefit of learning combination weights from the stronger claim that a complex ensemble consistently improves upon the best individual model.

\section{Related work}
\subsection{Solar representation and weather information}
Solar position provides a deterministic reference for PV forecasting. Established astronomical algorithms calculate elevation, azimuth, and seasonal geometry from location and time \cite{spa}, and pvlib makes these calculations available together with clear-sky irradiance models \cite{pvlib}. Such features describe illumination more directly than civil-clock position, particularly when sites differ in longitude. Their forecasting role is to organize the expected generation envelope; actual AC output also depends on atmospheric attenuation, array configuration, temperature, and plant operation.

The origin of weather information is equally relevant. An archived forecast represents a prediction for a future valid time, whereas an observation or reanalysis describes conditions with the benefit of later information. These sources support different experiments. The PVDAQ study uses the historical GFS forecast archive \cite{gfs}, preserving its native statistical intervals and declaring an initialization-based availability assumption. This approach allows comparison under a common information rule while keeping the distinction between a retrospective forecast experiment and a verified live service.

\subsection{Temporal models and forecast combination}
N-BEATS uses basis expansion and residual blocks to represent temporal patterns \cite{nbeats}. N-HiTS extends multiscale modeling through hierarchical interpolation and multirate processing \cite{nhits}, while TimeMixer decomposes and mixes information across temporal scales \cite{timemixer}. These designs motivate complementary temporal experts. Transformer alternatives organize the input differently: PatchTST represents subseries as patches \cite{patchtst}, iTransformer treats variates as tokens \cite{itransformer}, and TimeXer explicitly models endogenous and exogenous variables \cite{timexer}. They provide methodological context rather than unexecuted performance baselines in this study.

LightGBM is a particularly relevant comparator because lagged power, rolling statistics, solar features, and weather descriptors can be supplied directly to gradient-boosted trees \cite{lightgbm}. A neural ensemble should be assessed against this complete feature-based alternative rather than against a deliberately restricted tree model. Pretrained forecasting models introduce another useful comparison. Chronos and TimesFM study transfer from heterogeneous time series \cite{chronos1,timesfm}; Chronos-2 extends forecasting to multivariate inputs and future covariates \cite{chronos2}. We distinguish zero-shot prediction from station-specific fine-tuning because they involve different learning opportunities.

Forecast combination has a well-established statistical basis \cite{bates}. Stacked generalization and stacked regression learn how to combine model outputs \cite{wolpert,breiman}. The present method applies this principle through nonnegative, horizon-specific weights learned in chronological validation segments. Its contribution is not a new stacking algorithm. Rather, the experiments examine whether the chosen heterogeneous predictors contain useful complementary information and whether the resulting gain justifies retaining all components.

\subsection{Evaluation and evidence of generalization}
Shared energy forecasting tasks provide an important basis for reproducible comparison \cite{hong}. However, a public source containing many systems does not make an experiment on one selected station a multisite evaluation. Likewise, an hourly dataset cannot establish performance against native 15-minute targets. We retain the two resolutions and report the hourly adaptation as additional evidence about forecast combination, rather than as an exact replication of the solar-geometry model.

Forecast evaluation must also account for temporal dependence, scaling, and model selection \cite{hewamalage}. Conditions under which cross-validation is valid for autoregressive prediction do not justify arbitrary shuffling in a nonstationary operational setting \cite{bergmeir}. We separate training, early stopping, fusion fitting, and evaluation in time. Paired block resampling preserves dependence within consecutive dates \cite{kunsch}, while repeated seeds in the hourly experiment describe training variability. These checks address different sources of uncertainty and are reported separately.
\section{Forecasting task and input representation}
\subsection{Forecast targets and normalization}
Let $\Delta=15$ minutes, $t$ denote a forecast origin, and $H=16$. The target $P_{t,h}$ is the AC power measurement assigned to interval $[t+(h-1)\Delta,t+h\Delta)$, for $h=1,\ldots,H$. Horizon labels of 15--240 minutes refer to these interval endpoints. The selected meter timestamps are interpreted as interval starts in the research adapter. Instrument arrival times are not observed, so this is an explicit experimental convention, not a verified telemetry service guarantee.

We normalize by the public DC nameplate $C=893$ kW:
\begin{equation}
 y_{t,h}=P_{t,h}/C.\label{eq:normalization}
\end{equation}
The output is AC power, whereas the denominator is DC nameplate capacity. This distinction is retained throughout. Absolute MAE in kW is reported to make the practical error scale visible. The task predicts power; it does not directly score integrated energy or revenue.

The core historical context contains 96 intervals, or 24 hours, together with observation masks. Some engineered predictors also use historical aggregates, a 48-hour counterpart, or a longer causal analog bank. All measurements precede the origin. The comparison matches information sources and forecast availability, although effective historical receptive fields differ.

\subsection{Astronomical representation}
Solar geometry is calculated at each interval midpoint, using the public latitude $\phi=38.996306^{\circ}$, longitude $-122.134111^{\circ}$, and an assumed altitude of 10 m. Let $\omega$ be solar hour angle, $\delta$ declination, $\lambda$ geocentric solar longitude, $\alpha$ elevation, and $\psi$ azimuth. Their geometric relationship includes
\begin{equation}
 \sin\alpha=\sin\phi\sin\delta+\cos\phi\cos\delta\cos\omega.
\end{equation}
The 13-dimensional feature vector is
\begin{align}
 s_t=(&\sin\omega,\cos\omega,\sin\delta,\cos\delta,
 \sin\lambda,\cos\lambda,\sin\alpha,\nonumber\\
 &\sin\psi,\cos\psi,g^{\mathrm{cs}},d,d^{\uparrow},d^{\downarrow}),
\end{align}
where $g^{\mathrm{cs}}$ is Haurwitz clear-sky global horizontal irradiance divided by 1000 W m$^{-2}$, $d$ indicates positive solar elevation, and $d^{\uparrow},d^{\downarrow}$ mark grid transitions into and out of daylight. The annual phase is astronomical solar longitude, not a Gregorian day-of-year sine wave. Civil time is used for chronological split boundaries and reporting, not as a substitute for solar angle. This representation does not imply that every component is invariant to clock phase: the rolling analog candidates retain exact 24-hour lags.

\subsection{Forecast-weather representation and availability}
The GFS inputs comprise temperature, relative humidity, zonal and meridional wind, pressure, total cloud cover, downward shortwave radiation, and precipitation. Eight values, their eight native statistical-span descriptors, three forecast-cycle/lead/offset descriptors, and eight source-valid indicators produce 27 columns. Missing values remain masked. Neural weather means and scales are fitted only on the 2019--2022 training segment.

GFS fields have a $0.25^{\circ}$ spatial grid and native forecast statistics spanning three or six hours. In particular, an interval-average radiation or cloud forecast is not a new instantaneous observation every 15 minutes. Mapping it to the target grid preserves these descriptors. The research adapter assumes a forecast becomes available six hours after initialization. Actual historical reception times are unknown. Past weather at a context position is derived from the forecast available at that position; the future trajectory is derived from the forecast cycle available at the current prediction origin. Future power and future observed weather are excluded from the inference input.

\section{Hierarchical heterogeneous forecasting}
\begin{figure}[tbp]
\centering\includegraphics[width=\linewidth]{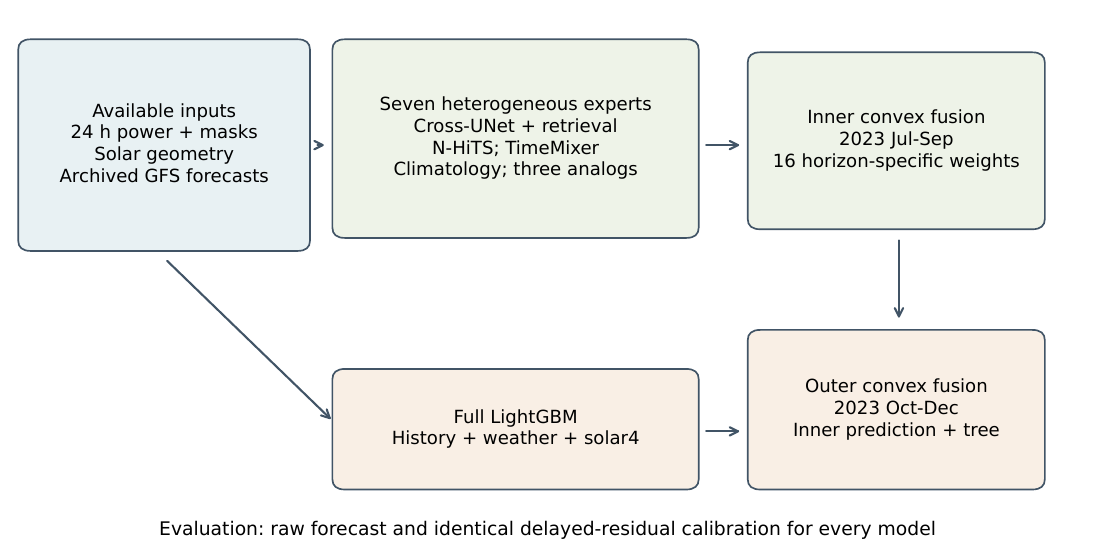}
\caption{Forecast construction and separate chronological fitting periods. Future weather is archived NWP available under the six-hour research rule. Delayed-residual calibration is applied equally to every evaluated model.}\label{fig:pipeline}
\end{figure}

Figure~\ref{fig:pipeline} summarizes the model. Each expert produces a vector in $\mathbb{R}^{16}$. We distinguish station-specific offline fitting, causal retrieval of previously observed power, horizon-specific fusion, and delayed error calibration.

\subsection{Neural experts and residual retrieval}
Three neural predictors are trained from fresh public-data initialization: an N-HiTS adaptation, a TimeMixer adaptation, and Cross-UNet. Historical power and masks are accompanied by historical and future solar and NWP features. Each minimizes the normalized absolute forecast loss,
\begin{equation}
 \mathcal{L}_{\mathrm{power}}=\frac{1}{NH}\sum_{t=1}^{N}\sum_{h=1}^{H}
 |y_{t,h}-f_{\theta,h}(\mathcal{I}_t)|.
\end{equation}
Cross-UNet uses 12-step segments, model width 256, four attention heads, three encoder scales, and feed-forward width 512. N-HiTS uses hidden width 512 and pooling scales 8, 4, and 1. TimeMixer uses model width 256, feed-forward width 512, three mixing layers, and two downsampling levels. These are the executed public adaptations rather than new standalone backbone algorithms.

The Cross-UNet forecast is augmented by a learned residual-retrieval gate. For a candidate historical window $j$, the stored target end must precede the query origin by at least 120 minutes. Candidates belong to the same station and the frozen training bank; their solar-hour slot is matched, with a 45-bin seasonal tolerance in a 366-bin solar-longitude cycle and a configured seasonal fallback. At least 48 common observed history points are required. Distances combine power, weather, and solar terms with weights 0.4, 0.5, and 0.1. Within weather matching, radiation and cloud variables receive weights 2.0 and 1.5, while the other columns receive 0.5.

Up to 32 neighbors produce a weighted residual estimate $\bar r_{t,h}$ and statistics of dispersion, distance, coverage, and effective neighbor count. A small horizon-aware gate forms
\begin{equation}
 \hat y^{\mathrm{CR}}_{t,h}=\max\{0,\hat y^{\mathrm{Cross}}_{t,h}
       +a_{t,h}\bar r_{t,h}\},\qquad 0\leq a_{t,h}\leq1.5.
\end{equation}
Its hidden width is 32, and station and horizon embeddings each have width 4. The gate is trained for 50 epochs on the covered training windows. Training-bank residuals are in-sample residuals of the fitted Cross-UNet; they are not forward out-of-fold residuals. This may affect gate generalization, although evaluation targets are excluded from both backbone and gate fitting.

\subsection{State climatology and rolling analog experts}
A train-only climatology $c(\omega)$ averages power by solar-hour slot. At prediction time, the most recent four intervals determine a multiplicative state ratio using valid observations and reference power above 1\% of capacity. The ratio is clipped to $[0,2]$ and applied to the future climatological curve. If the ratio cannot be estimated, the scale defaults to one. The forecast is bounded to $[0,1.25C]$.

Three analog experts use $(K,D,M)=(4,45,60)$, $(8,180,360)$, and $(8,90,180)$, where $K$ is the maximum number of neighbors, $D$ the lookback in days, and $M$ the history-match duration in minutes. Candidate origins are at exact integer-day offsets. Their full history and forecast target must already be observed at the query time. With normalized historical vectors $x_t$ and $x_j$, their score is
\begin{equation}
 d_{tj}=\frac{1}{L}\|x_t-x_j\|_1
        +0.15|\operatorname{std}(x_t)-\operatorname{std}(x_j)|.
\end{equation}
Selected neighbors receive weights proportional to
\begin{equation}
 w_{tj}\propto(d_{tj}+0.01)^{-2}\exp(-\mathrm{age}_{tj}/D).
\end{equation}
Each retrieved future curve is adjusted by the ratio of current to candidate mean history, clipped to $[0.55,1.6]$. A state-climatology fallback handles unavailable candidates. These analog banks roll forward causally and may incorporate already observed evaluation-period history; they never include power that has not matured at the origin. This is an online-history protocol, not a permanently frozen training-history-only comparison.

\subsection{Full LightGBM and solar-scaled persistence features}
The eighth expert is a separate LightGBM regressor for each horizon. It retains the full feature builder, including recent lags, rolling summaries, availability masks, solar features, and weather descriptors. Each horizon receives 311 columns: 170 shared, 137 horizon-specific, and four solar-derived features. Unavailable weather metadata and five-minute power detail remain explicitly missing.

Four solar-derived features encode recent power relative to clear-sky irradiance and solar-scaled persistence, with validity flags. Let $p_-$ be the latest normalized observed power and $g_-^{\mathrm{cs}}$ its dimensionless clear-sky irradiance. If the power is valid and $g_-^{\mathrm{cs}}\geq0.05$, define
\begin{equation}
 q_t=p_-/g_-^{\mathrm{cs}},\qquad
 \hat p^{\mathrm{solar}}_{t,h}=q_t g_{t,h}^{\mathrm{cs}}.
\end{equation}
Invalid ratios remain missing. Because the numerator is normalized AC power and the denominator is a clear-sky horizontal-irradiance proxy, $q_t$ is an empirical index, not panel conversion efficiency. Forecast radiation and clouds are separate tree inputs; the model can learn their conditional association with power without assuming a universal linear conversion law.

\subsection{Chronological two-level convex fusion}
For horizon $h$, let $p_{t,h,k}$ denote the first seven experts. On the first fusion-validation segment $V_1$, inner weights solve
\begin{equation}
 \hat u_h=\arg\min_{u\in\Delta^7}\frac{1}{|V_{1,h}|}
 \sum_{t\in V_{1,h}}\left|y_{t,h}-\sum_{k=1}^{7}u_kp_{t,h,k}\right|,
\end{equation}
where $\Delta^K=\{w\in\mathbb{R}^{K}:w\geq0,\sum_kw_k=1\}$ and $V_{1,h}$ contains daylight targets. Define $z_{t,h}=\sum_k\hat u_{h,k}p_{t,h,k}$. On a later validation segment $V_2$, outer weights solve
\begin{equation}
 \hat v_h=\arg\min_{v\in\Delta^2}\frac{1}{|V_{2,h}|}
 \sum_{t\in V_{2,h}}|y_{t,h}-v_1z_{t,h}-v_2p^{\mathrm{LGB}}_{t,h}|.
\end{equation}
The final raw forecast is $\hat y_{t,h}=\hat v_{h,1}z_{t,h}+\hat v_{h,2}p^{\mathrm{LGB}}_{t,h}$. Both optimizations are linear programs with absolute-error slack variables. We fit 32 programs in total and freeze all weights before scoring the evaluation period. Weights vary by horizon, but are fixed across forecast origins: this experiment does not implement weather-dependent dynamic gating at the fusion level.

\subsection{Delayed-residual calibration}
The separate calibration policy, denoted state50 in the implementation, is identical for all models. For each horizon it retains at most 384 matured residuals within 30 days and requires at least 64 before calibration. A target residual becomes eligible only after its interval endpoint plus an additional 60-minute delay. Define $e_{j,h}=y_{j,h}-\hat y_{j,h}$ and
\begin{equation}
 b_{t,h}=\clip\left(\med_{j\in\mathcal{H}_{t,h}}e_{j,h},-0.15,0.15\right).
\end{equation}
Predictions are assigned to three power bins with thresholds 0.15 and 0.50 per unit. If the current bin also contains at least 64 residuals, $b_{t,h}$ is replaced by the average of the clipped global and bin-specific medians. The corrected forecast is
\begin{equation}
 \tilde y_{t,h}=\clip(\hat y_{t,h}+0.5b_{t,h},0,1.25).
\end{equation}
Before sufficient residual history exists, the raw prediction is retained. Every model has its own residual history and begins evaluation with an empty calibrator. Residuals are computed against the raw prediction, not recursively against a previously corrected prediction.

\section{Experimental design}
\subsection{Dataset, quality control, and chronological splits}
PVDAQ station 2107 is the public Farm Solar Array in Arbuckle, California \cite{pvdaq}. We use its revenue-grade AC output meter, identified as {\small\texttt{meter\_revenue\_grade\_ac\_output\_meter\_149578}}. The selected signal is natively on a 15-minute grid. The source audit found no off-grid records for this meter.

The measurement grid contains finite, nonnegative observations as valid power. A forecast window is eligible when all 16 targets are valid, the latest historical interval is valid, and at least 48 of the 96 historical intervals are observed. Windows crossing a split endpoint are excluded. Neither irradiance availability nor weather download completeness determines the power evaluation cohort. We do not smooth low-cloud V- or U-shaped power excursions or alter public evaluation labels. Any residual undetected operational anomalies remain a limitation of the measured data.

\begin{table}[tbp]\centering
\caption{Chronological public-data partition. Calendar boundaries use America/Los\_Angeles. Windows contain all sixteen future targets.}\label{tab:splits}
\begin{tabularx}{\linewidth}{l>{\raggedright\arraybackslash}Xr}\toprule Segment & Purpose & Windows\\\midrule
2019--2022 & Fit backbones, transforms, and training banks & 139,869\\
2023 Jan--Jun & LightGBM early stopping & 17,197\\
2023 Jul--Dec & Fusion fitting and single-model selection & 12,636\\
2024 Jan--Oct & Public development evaluation & 29,211\\\bottomrule
\end{tabularx}
\end{table}
\FloatBarrier
Inner fusion uses July--September 2023 and outer fusion uses October--December 2023. Removing windows that cross the internal boundary leaves 8,816 and 3,804 fitting windows, respectively. The strongest standalone supervised comparator is selected using October--December daylight horizon-averaged error; this selects LightGBM. The 2024 period had been exposed in earlier public-data work and is therefore explicitly a development replication, not a newly untouched test set.

The cohort requires 33,252 archived GFS source keys, of which 33,239 were decoded and verified. Thirteen unavailable endpoints retain their associated power windows with masked weather. PVDAQ uses GFS only; dual-source weather experiments are outside the present comparison.

\subsection{Training budgets and comparator inputs}
The three neural backbones each receive 12 epochs using AdamW, learning rate $3\times10^{-4}$, weight decay $10^{-4}$, batch size 256, and gradient-norm clipping at 1. Their parameter counts are approximately 2.27 million (N-HiTS), 3.51 million (TimeMixer), and 6.90 million (Cross-UNet). The base initialization seed is 2021. TimeMixer additionally receives 2,400 low-learning-rate updates at $3\times10^{-6}$, with its fixed continuation seed 20260909. Thus, ``single seed'' here refers to one fixed base-seed configuration, not identical random streams in every subroutine.

Each LightGBM uses an L1 objective, learning rate 0.03, 63 leaves, minimum leaf count 100, feature and bagging fractions 0.85, L1/L2 regularization 0.1/1.0, and at most 1,800 rounds with 120-round early stopping. The same public training windows and declared feature schema are retained; a small surrogate tree does not substitute for this full baseline.

Chronos-2 is evaluated both without station-specific fine-tuning and after 3,000 full-parameter optimizer updates at learning rate $10^{-6}$, using its native quantile loss. Its input includes the same 24-hour power context, 13 solar and 27 GFS features, and a 48-hour power counterpart. Each training task uses 112 positions split into 96 context and 16 forecast positions. Future covariates come from the task's own forecast origin. Future power labels and cross-task learning are excluded at inference.

Each fine-tuning update contains 25 station windows represented by 1,050 component series, accumulated over five microbatches. The run uses approximately 75,000 sampled task exposures. The official pretrained snapshot is fixed, and no evaluation score selects a checkpoint. Pretraining data overlap with this public station is not independently known. The neural and foundation-model budgets are deliberately disclosed; they are not FLOP-matched.

\subsection{Metrics, uncertainty, and decision rule}
Let $\mathcal{D}_h$ be eligible evaluation targets with solar elevation above $5^{\circ}$ at their interval midpoints. The primary metric is
\begin{equation}
 E=\frac{1}{H}\sum_{h=1}^{H}\frac{1}{|\mathcal{D}_h|}
 \sum_{t\in\mathcal{D}_h}|\tilde y_{t,h}-y_{t,h}|.
\end{equation}
This horizon-macro nMAE gives each forecast horizon equal weight. We also report its score form $S=100(1-E)$, solely to connect with the predeclared decision rule. It is not classification accuracy. A score difference is expressed in percentage points (pp), whereas relative MAE reduction is $(E_{\mathrm{control}}-E_{\mathrm{ensemble}})/E_{\mathrm{control}}$. RMSE is computed from horizon-averaged mean squared error. Daylight and all-times metrics use separate, common masks. All sixteen horizons and the 240-minute endpoint are reported.

Paired uncertainty is estimated with 2,000 circular seven-day block resamples and fixed bootstrap seed 2021. A common date draw is applied to both models, preserving within-day dependence and recalculating per-horizon denominators. Intervals are percentile 95\% intervals conditional on the fixed station, evaluation period, and fitted models. They do not incorporate retraining uncertainty, public-period exposure, or uncertainty across sites. Because these are unadjusted intervals for the declared comparisons, they should not be read as a general familywise significance guarantee.

Before scoring, the practical rule required an all-horizon score gain of at least 0.2 pp with a positive lower interval endpoint against both the validation-selected strongest single model and fine-tuned Chronos-2, and no degradation in the 240-minute point estimate. A statistically positive effect that falls below this practical threshold is reported without changing the threshold after seeing results.

\subsection{Expert removal and hourly adaptation}
The PVDAQ removal experiment excludes one expert at a time and refits the remaining fusion weights using the same 2023 validation segments. Backbone predictions, evaluation targets, and calibration rules are unchanged. This isolates each expert's contribution within the specified combination procedure rather than measuring the effect of changing its training budget. Eight removal comparisons require 240 additional linear programs. We report both nominal paired intervals and 99.375\% percentile intervals using a Bonferroni adjustment for eight comparisons.

GEFCom2014 Solar provides an independent public source with three anonymous zones and hourly power and weather predictors \cite{hong}. The adaptation predicts the next one to four hours using a 24-hour neural context. Training covers April 2012--September 2013; October--December 2013 is used for selection, January--May 2014 for fusion, and June 2014 for evaluation. These partitions contain 34,458, 5,793, 9,510, and 1,887 windows, respectively. All zones occur in training, so the experiment evaluates future periods at known zones.

The hourly adapter uses 12 ECMWF forecast columns. Accumulated radiation is differenced within each daily forecast cycle, and the power column in the predictor files is excluded from weather inputs. A cycle is assumed available at midnight, and each four-hour trajectory is restricted to that cycle. Published timestamps are retained without an inferred time-zone conversion. Coordinates are anonymous, so calendar features replace astronomical coordinates; no site-specific solar angles are inferred. This is a resolution- and input-adapted version of the ensemble.

The hourly experiment uses three seeds, 2021--2023. For each seed, the three neural backbones receive 12 epochs, TimeMixer receives 2,400 additional low-learning-rate updates, four horizon-specific LightGBM models use the same 1,800-round ceiling, and the retrieval gate receives 50 epochs. Fusion requires eight linear programs per seed. Inner fusion uses January--March 2014 and outer fusion uses April--May. The hourly analog settings $(K,D,M)$ are $(4,45,4)$, $(8,180,24)$, and $(8,90,12)$, with $M$ now measured in hours; they use inverse-square similarity weighting with an age penalty. All models share forecast origins and target availability, although their memory structures differ.

The primary hourly metric is all-times normalized MAE averaged over the four horizons, with the four-hour endpoint also reported. It is not pooled with PVDAQ daylight error. Paired resampling uses 2,000 circular seven-day block draws, shared across zones and seeds, preserving calendar dependence. Seed standard deviations characterize training variability, whereas the block intervals characterize variation over the evaluated dates.
\section{Results and discussion}
\subsection{Aggregate and horizon-specific performance}
\begin{table}[tbp]\centering\small
\caption{Public PVDAQ2107 daylight results with common delayed-residual calibration. Lower is better. All-horizon columns average the sixteen horizons; the final column is the 240-minute endpoint.}\label{tab:main}
\begin{tabular}{lrrrr}\toprule Model & nMAE (\%) & MAE (kW) & RMSE (kW) & 4 h nMAE (\%)\\\midrule
Hierarchical eight-expert & 4.315 & 38.535 & 72.825 & 5.243 \\
Full LightGBM & 4.500 & 40.189 & 74.624 & 5.422 \\
Chronos-2 fine-tuned & 4.592 & 41.008 & 82.979 & 5.663 \\
N-HiTS adaptation & 4.763 & 42.529 & 77.690 & 5.857 \\
TimeMixer + lowLR & 4.936 & 44.075 & 84.698 & 5.898 \\
Cross-UNet + retrieval & 5.248 & 46.864 & 81.828 & 6.215 \\
Cross-UNet base & 6.977 & 62.302 & 92.735 & 8.881 \\
Equal eight-expert & 5.234 & 46.738 & 80.368 & 7.014 \\
Chronos-2 zero-shot & 7.701 & 68.766 & 121.863 & 11.268 \\
Solar persistence & 10.101 & 90.198 & 147.548 & 17.798 \\
Persistence & 19.508 & 174.204 & 228.911 & 31.400 \\
\bottomrule
\end{tabular}

\end{table}
\FloatBarrier
The hierarchical ensemble attains daylight nMAE 4.315243\%, compared with 4.500400\% for full LightGBM and 4.592130\% for fine-tuned Chronos-2 (Table~\ref{tab:main}). Absolute MAE decreases from 40.189 to 38.535 kW relative to LightGBM and from 41.008 to 38.535 kW relative to Chronos-2. These are relative MAE reductions of 4.11\% and 6.03\%. The corresponding score gains are 0.185156 and 0.276886 pp.

\begin{figure}[tbp]\centering
\includegraphics[width=\linewidth]{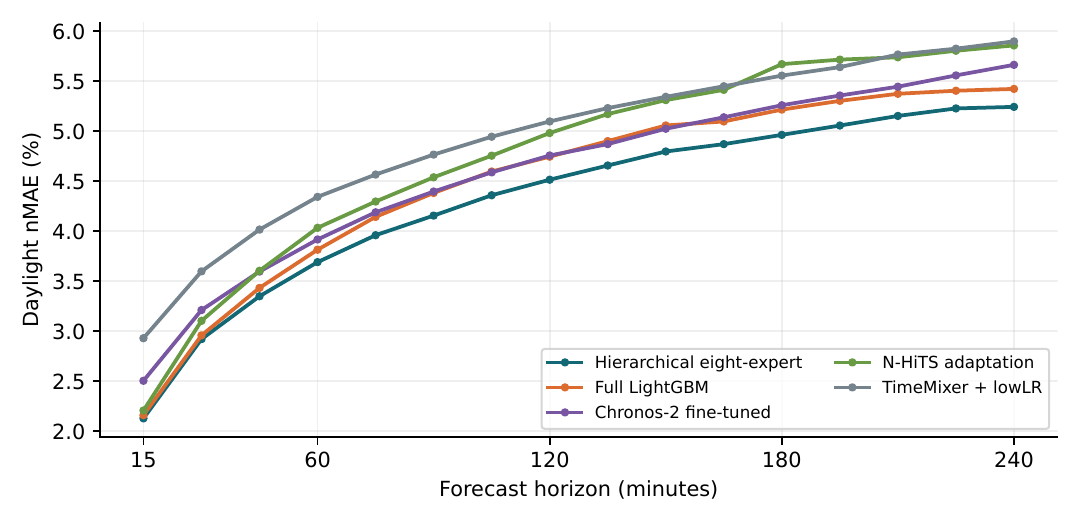}
\caption{Daylight nMAE at all sixteen horizons under identical delayed-residual calibration. Curves use the same evaluation keys and weather source.}\label{fig:horizons}
\end{figure}

At 240 minutes, ensemble nMAE is 5.242859\%, compared with 5.422484\% for LightGBM and 5.663343\% for fine-tuned Chronos-2. Figure~\ref{fig:horizons} and Appendix~\ref{app:horizons} show the complete horizon profile; the aggregate gain is not inferred from only a favorable short horizon. Zero-shot Chronos-2 performs less strongly on this station, while full fine-tuning closes much of that gap. This contrast supports reporting adaptation status explicitly, rather than using zero-shot performance alone as evidence against foundation models.

\subsection{Paired uncertainty and practical magnitude}
\begin{table}[tbp]\centering\small
\caption{Paired score improvement of the ensemble, with seven-day block percentile intervals. Positive values favor the ensemble. These are nominal, conditional intervals.}\label{tab:ci}
\begin{tabular}{llrr}\toprule Control & Horizon & Gain (pp) & 95\% interval (pp)\\\midrule
Full LightGBM & All 16 & +0.1852 & [+0.1181, +0.2470] \\
Full LightGBM & 240 min & +0.1796 & [+0.1059, +0.2512] \\
Chronos-2 fine-tuned & All 16 & +0.2769 & [+0.0669, +0.5051] \\
Chronos-2 fine-tuned & 240 min & +0.4205 & [+0.0345, +0.8304] \\
\bottomrule
\end{tabular}

\end{table}
\FloatBarrier
Table~\ref{tab:ci} shows positive interval lower endpoints for both declared controls at all horizons combined and at 240 minutes. The LightGBM all-horizon gain is 0.1852 pp with interval $[0.1181,0.2470]$ pp; the Chronos-2 fine-tuned gain is 0.2769 pp with interval $[0.0669,0.5051]$ pp. The evidence supports positive conditional aggregate differences.

Nevertheless, the gain over LightGBM is below the predeclared 0.2 pp practical threshold. The estimated improvement is therefore statistically positive under the resampling protocol but smaller than the specified practical target. An application may weigh a 1.65 kW average absolute-error reduction differently depending on computing cost and operational value, which require evidence beyond a percentage score.

\subsection{Learned fusion and effective expert weights}
The equal-weight eight-expert forecast has daylight nMAE 5.233791\%, substantially above the hierarchical result. This provides a direct control for the choice of weights while keeping the same expert forecasts. The complementary removal experiment below evaluates individual experts after refitting the validation weights.

\begin{figure}[tbp]\centering
\includegraphics[width=\linewidth]{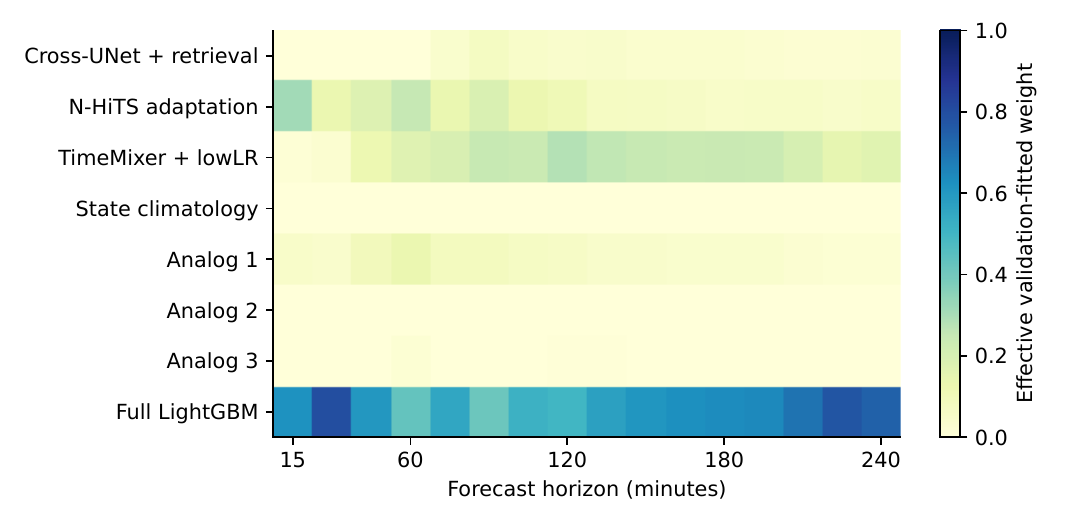}
\caption{Effective validation-fitted weights. Sparse weights highlight component redundancy at some horizons. No weights are fitted on the 2024 evaluation scores.}\label{fig:weights}
\end{figure}

The effective weights are $\hat v_{h,1}\hat u_{h,k}$ for the first seven experts and $\hat v_{h,2}$ for LightGBM (Figure~\ref{fig:weights}). LightGBM receives approximately 62.0\% of the weight at 15 minutes and 73.9\% at 240 minutes. N-HiTS receives approximately 31.4\% at 15 minutes, while TimeMixer receives approximately 16.6\% at 240 minutes. Some components have zero or very small weights at individual horizons. The public station therefore does not support the assertion that all eight experts are necessary, or a universal rule that tree weight must decrease with lead time. These weights are fitted validation coefficients, not causal feature attributions.

\subsection{Calibration and monthly stability}
\begin{table}[tbp]\centering\small
\caption{Calibration comparison on the same daylight targets. Gain is calibrated minus raw score; a negative value means calibration hurts.}\label{tab:calibration}
\begin{tabular}{lrrr}\toprule Model & Raw nMAE (\%) & Calibrated nMAE (\%) & Gain (pp)\\\midrule
Hierarchical eight-expert & 4.3484 & 4.3152 & +0.0332 \\
Full LightGBM & 4.5405 & 4.5004 & +0.0401 \\
Chronos-2 fine-tuned & 4.5916 & 4.5921 & -0.0005 \\
Equal eight-expert & 5.3315 & 5.2338 & +0.0977 \\
Chronos-2 zero-shot & 7.8866 & 7.7006 & +0.1860 \\
\bottomrule
\end{tabular}

\end{table}
\FloatBarrier
Calibration improves the ensemble score by 0.0332 pp and LightGBM by 0.0401 pp, whereas the effect on fine-tuned Chronos-2 is slightly negative ($-0.0005$ pp). The raw ensemble already outperforms both controls; its advantage is not created solely by providing a correction unavailable to competitors. Calibration contributes a small, model-dependent change and should not be assumed beneficial in every setting.

\begin{figure}[tbp]\centering
\includegraphics[width=\linewidth]{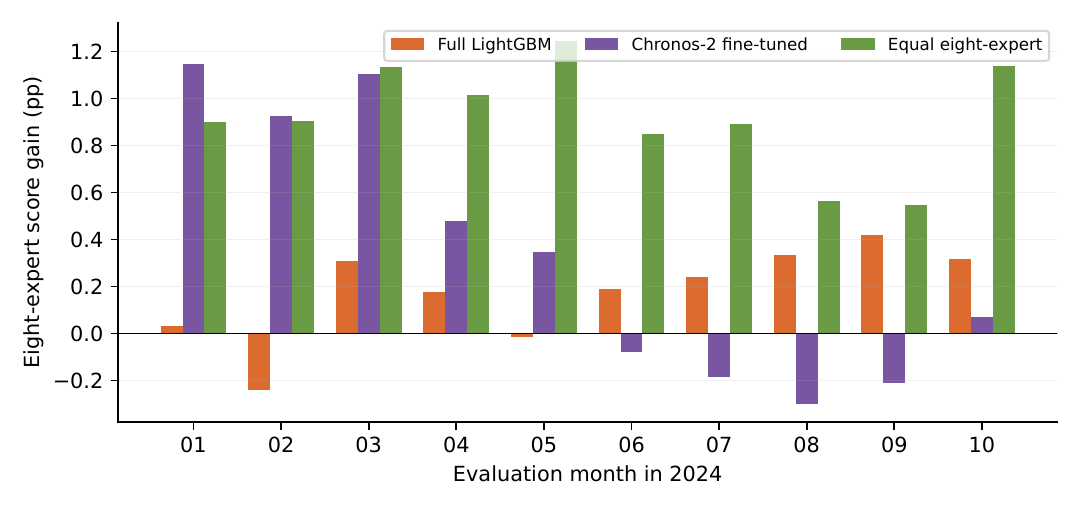}
\caption{Monthly score gains under common calibration. Negative months are retained. The aggregate comparison does not imply dominance in each season.}\label{fig:months}
\end{figure}

Figure~\ref{fig:months} and Appendix~\ref{app:months} report all ten calendar months. The ensemble exceeds LightGBM in eight months and fine-tuned Chronos-2 in six. Monthly gains range from $-0.2418$ to $+0.4195$ pp against LightGBM and from $-0.3000$ to $+1.1477$ pp against fine-tuned Chronos-2. In contrast, learned fusion exceeds equal weighting in every month, by $+0.5480$ to $+1.2459$ pp. These are post hoc descriptive summaries of an already exposed evaluation period, not independent monthly hypothesis tests or a rule for selecting favorable months.

\subsection{Contribution and redundancy of individual experts}
Table~\ref{tab:remove} reports the change in error after each expert is removed and the remaining weights are refitted. Removing N-HiTS, TimeMixer, or LightGBM increases the aggregate point estimate of error. The largest point difference follows removal of LightGBM, consistent with its substantial fitted weight. However, the uncertainty intervals differ, and only the N-HiTS contribution remains strictly positive after adjustment for all eight comparisons.
\begin{table}[tbp]\centering\small
\caption{PVDAQ expert removal under common calibration. Positive differences indicate lower error for the complete ensemble. Intervals in the final column account for eight comparisons. Units are percentage points of normalized MAE.}\label{tab:remove}
\begin{tabular}{lrrr}\toprule
Removed expert & Difference & Nominal 95\% interval & Adjusted interval\\\midrule
Cross-UNet + retrieval & 0.0043 & [0.0001, 0.0084] & [$-0.0014$, 0.0098]\\
N-HiTS & 0.0376 & [0.0228, 0.0527] & [0.0178, 0.0578]\\
TimeMixer & 0.0642 & [0.0074, 0.1166] & [$-0.0199$, 0.1341]\\
State climatology & 0.0000 & [0.0000, 0.0000] & [0.0000, 0.0000]\\
Analog 1 & $-0.0012$ & [$-0.0064$, 0.0036] & [$-0.0083$, 0.0051]\\
Analog 2 & 0.0000 & [0.0000, 0.0000] & [0.0000, 0.0000]\\
Analog 3 & $-0.0001$ & [$-0.0004$, 0.0002] & [$-0.0005$, 0.0004]\\
LightGBM & 0.1390 & [0.0189, 0.2625] & [$-0.0267$, 0.3189]\\\bottomrule
\end{tabular}
\end{table}
The state climatology and second analog expert can be removed without changing the reported aggregate result. The other analog removals provide no reliable evidence of a positive contribution. Thus, eight predictors describe the evaluated configuration, not a demonstrated optimum in model count. A smaller ensemble is therefore worth investigating through a new independent evaluation.

\subsection{Results on GEFCom2014 Solar}
The three-seed hourly experiment yields mean normalized MAE of 2.140543\% for the ensemble and 2.142789\% for LightGBM (Table~\ref{tab:gefcom}). The difference is only 0.002245 percentage points, with a paired interval of [$-0.035991$, 0.030584]. At four hours, the point estimate slightly favors the selected single model. The experiment therefore provides no clear evidence that the ensemble improves upon LightGBM in this setting.
\begin{table}[tbp]\centering\small
\caption{GEFCom2014 Solar: all-times normalized MAE (\%) averaged over native one- to four-hour horizons. Values are means and sample standard deviations across three seeds; lower is better. Raw forecasts are used throughout this table.}\label{tab:gefcom}
\begin{tabular}{lrr}\toprule Model & Mean & Seed standard deviation\\\midrule
Hierarchical eight-expert & 2.140543 & 0.007563\\
Full LightGBM & 2.142789 & 0.004861\\
N-HiTS adaptation & 2.587379 & 0.041635\\
TimeMixer + lowLR & 2.388756 & 0.033634\\
Cross-UNet + retrieval & 2.662501 & 0.159178\\
Equal eight-expert & 2.409170 & 0.018927\\
State climatology & 3.512451 & 0.000000\\
Persistence & 7.107903 & 0.000000\\
Previous-day persistence & 3.748680 & 0.000000\\\bottomrule
\end{tabular}
\end{table}
Learned fusion nevertheless improves upon equal weighting by 0.268626 percentage points, with interval [0.093798, 0.454062]. The corresponding four-hour improvement is 0.334148 percentage points, with interval [0.151189, 0.521688]. This agrees with the PVDAQ finding that weighting matters, while showing that improved weighting does not necessarily create a substantial advantage over a strong tree predictor. The anonymous locations, hourly sampling, and one-month evaluation period limit direct comparison with the solar-aware PVDAQ experiment.

\subsection{Computational requirements and numerical consistency}
Component-level timing was performed on an NVIDIA RTX A6000 using 128 fixed evaluation origins, one warm-up run, and the median of three measured runs. The 16 LightGBM predictors require 0.338 ms per origin after feature construction; their feature builder adds 0.060 ms. Batched N-HiTS, TimeMixer, and Cross-UNet inference requires 0.041, 0.133, and 0.182 ms per origin, respectively. At batch size one, these neural costs increase to 4.324, 15.449, and 19.253 ms. Climatology and analog retrieval together require 2.425 ms, and residual-neighbor search adds 1.252 ms per origin. The complete component table accompanies the manuscript.

These measurements locate the principal computational costs but are not an end-to-end service benchmark. They exclude model loading, weather acquisition, raw-file parsing, and residual-history maintenance. Chronos-2 timing includes its native pipeline preparation, whereas the neural component timings use GPU-resident inputs. Consequently, those timings have different boundaries and should not be interpreted as equal-scope deployment comparisons. Nor does timing alone equalize the total training resources available to each method.

Saved-model replay and independent metric recalculation support numerical consistency. A batch-size sensitivity analysis recomputed all 29,211 PVDAQ windows at batch size 128 with unchanged fusion weights. The aggregate raw ensemble score changed by approximately 0.000009 percentage points. Individual predictions, particularly those from Cross-UNet, were more batch-sensitive, so the scoring batch and floating-point settings remain part of the reproducible configuration. All reported primary scores retain their original predictions.

\subsection{Interpretation and practical implications}
Across the two public sources, the most consistent result is the advantage of learned fusion over equal weighting. The improvement over LightGBM is more conditional: it is positive on PVDAQ but negligible on the hourly GEFCom task. This distinction matters when choosing a forecasting system. A strong feature-based predictor can capture much of the dependence between recent power, solar position, and weather. Additional experts are useful only insofar as their errors provide complementary information that the validation procedure can identify.

The removal results sharpen this interpretation. Model diversity by construction does not guarantee a distinct contribution after fitting. Some historical components receive little effective weight or can be removed without a measurable loss. Conversely, the positive adjusted N-HiTS removal interval indicates that at least one temporal predictor adds information within this configuration. The results favor measuring complementarity directly rather than treating model count as evidence of forecasting quality.

For energy management, a reduction in power error may improve the input to storage scheduling or reserve allocation. Those downstream benefits are not quantified here. On PVDAQ, the gain over LightGBM corresponds to about 1.65 kW in average absolute error for an 893 kW DC-nameplate system. Whether this reduction justifies additional model maintenance depends on the operating objective and the cost of forecast errors. The timing results and observed component redundancy provide a basis for evaluating that trade-off in a subsequent operational study.

\subsection{Limitations}
Several limitations affect the scope of inference. PVDAQ includes one station and one fixed base-seed configuration. GEFCom adds three seeds and three known zones, but uses anonymous coordinates, hourly targets, and only June 2014 for evaluation. Neither experiment establishes transfer to an unseen plant. The public evaluation periods have also been exposed during development; chronological fitting in the reported runs does not eliminate possible influence from earlier design decisions. The date-block intervals are therefore conditional uncertainty summaries, not a substitute for a new blind evaluation.

The weather experiments rely on explicit availability assumptions. Initialization plus six hours for GFS and daily midnight for the GEFCom forecast cycle do not constitute historical reception records. Coarse cloud and radiation statistics also cannot resolve every local power fluctuation. Separate refitting ablations are needed to isolate the contribution of solar geometry, cloud forecasts, and radiation forecasts. In particular, a low cloud forecast is insufficient evidence that a short measured power valley is erroneous, and evaluation labels are not smoothed using weather.

Finally, the retrieval gate uses training-bank residuals rather than forward out-of-fold residuals. The rolling analogs and calibration also update from already matured observations during evaluation. Their causal use is explicit, but their memory differs from that of a predictor restricted to a fixed historical context. Future work should examine forward residual fitting, reduced expert sets, matched computational budgets, and longer multisite evaluations with verified forecast reception times.

\section{Conclusions}
This study evaluated horizon-specific hierarchical fusion for photovoltaic power forecasting using solar representations, archived weather forecasts, temporal models, and historical predictors. On PVDAQ station 2107, the ensemble attained daylight normalized MAE of 4.315\% over 15--240-minute horizons, with relative error reductions of 4.11\% against full-feature LightGBM and 6.03\% against fine-tuned Chronos-2 under common calibration. The gain over LightGBM was positive under paired resampling but remained below the predefined practical threshold.

The GEFCom2014 adaptation improved upon equal weighting while performing comparably to LightGBM across three training seeds. Expert removal further showed that several components were redundant. Together, these results support learning horizon-specific combinations, while indicating that the value of a larger ensemble depends on its measurable complementarity. A reduced configuration and prospective multisite evaluation are the next steps toward assessing operational usefulness.

\section*{Data availability}
PVDAQ measurements and metadata are available from the Open Energy Data Initiative \cite{pvdaq}; historical GFS forecasts are available from the NCAR Geoscience Data Exchange \cite{gfs}. GEFCom2014 Solar is described by Hong et al.\ \cite{hong}. Derived metric tables, removal contrasts, configurations, and timing results accompany this source package. A public implementation repository remains to be established.


\FloatBarrier

\appendix
\numberwithin{table}{section}
\numberwithin{figure}{section}
\section{Complete horizon results}\label{app:horizons}
\begin{table}[!ht]\centering\small
\caption{Daylight nMAE (\%) by horizon with common calibration. FT denotes full-parameter fine-tuning. These are the same predictions as Table~\ref{tab:main}.}
\begin{tabular}{rrrrrr}\toprule Minutes & Eight & LightGBM & Chronos-2 FT & N-HiTS & TimeMixer\\\midrule
15 & 2.128 & 2.160 & 2.504 & 2.207 & 2.930 \\
30 & 2.921 & 2.958 & 3.211 & 3.104 & 3.598 \\
45 & 3.349 & 3.433 & 3.596 & 3.604 & 4.016 \\
60 & 3.691 & 3.815 & 3.917 & 4.034 & 4.343 \\
75 & 3.960 & 4.143 & 4.189 & 4.296 & 4.566 \\
90 & 4.156 & 4.382 & 4.396 & 4.539 & 4.766 \\
105 & 4.359 & 4.596 & 4.588 & 4.755 & 4.945 \\
120 & 4.515 & 4.745 & 4.757 & 4.982 & 5.097 \\
135 & 4.657 & 4.900 & 4.871 & 5.171 & 5.230 \\
150 & 4.797 & 5.057 & 5.025 & 5.310 & 5.343 \\
165 & 4.870 & 5.097 & 5.139 & 5.413 & 5.449 \\
180 & 4.963 & 5.216 & 5.259 & 5.670 & 5.555 \\
195 & 5.056 & 5.303 & 5.357 & 5.715 & 5.641 \\
210 & 5.152 & 5.374 & 5.445 & 5.739 & 5.766 \\
225 & 5.227 & 5.404 & 5.557 & 5.804 & 5.825 \\
240 & 5.243 & 5.422 & 5.663 & 5.857 & 5.898 \\
\bottomrule
\end{tabular}

\end{table}

\begin{table}[!ht]\centering\small
\caption{All-horizon daylight and all-times nMAE. Nighttime targets make all-times error smaller; this does not constitute a separate improvement of the model.}
\begin{tabular}{lrr}\toprule Model & Daylight nMAE (\%) & All-times nMAE (\%)\\\midrule
Hierarchical eight-expert & 4.3152 & 2.1168 \\
Full LightGBM & 4.5004 & 2.1971 \\
Chronos-2 fine-tuned & 4.5921 & 2.2892 \\
Equal eight-expert & 5.2338 & 2.5881 \\
\bottomrule
\end{tabular}

\end{table}

\Needspace{18\baselineskip}
\section{Monthly differences}\label{app:months}
\begin{table}[!ht]\centering\small
\caption{Ensemble score minus comparator score, in percentage points, for every evaluation month. The month table is descriptive and uses the same common calibration.}
\begin{tabular}{lrrr}\toprule Month & vs. LightGBM & vs. Chronos-2 FT & vs. equal weights\\\midrule
2024-01 & +0.0312 & +1.1477 & +0.8997 \\
2024-02 & -0.2418 & +0.9248 & +0.9057 \\
2024-03 & +0.3095 & +1.1051 & +1.1338 \\
2024-04 & +0.1759 & +0.4808 & +1.0163 \\
2024-05 & -0.0129 & +0.3453 & +1.2459 \\
2024-06 & +0.1885 & -0.0801 & +0.8508 \\
2024-07 & +0.2385 & -0.1856 & +0.8916 \\
2024-08 & +0.3340 & -0.3000 & +0.5647 \\
2024-09 & +0.4195 & -0.2103 & +0.5480 \\
2024-10 & +0.3170 & +0.0725 & +1.1381 \\
\bottomrule
\end{tabular}

\end{table}

\FloatBarrier
\bibliographystyle{elsarticle-num}
\bibliography{references}
\end{document}